\documentclass[acmtog, nonacm]{acmart} % ,anonymous,
\usepackage{booktabs} % For formal tables
\usepackage{amsmath}
\usepackage{multirow}
\usepackage{color}
\DeclareMathOperator*{\argmin}{arg\,min}

\usepackage[capitalize]{cleveref}
\crefname{section}{Sec.}{Secs.}
\Crefname{section}{Section}{Sections}
\Crefname{table}{Table}{Tables}
\crefname{table}{Tab.}{Tabs.}

\usepackage[ruled]{algorithm2e} % For algorithms

\SetAlFnt{\small}
\SetAlCapFnt{\small}
\SetAlCapNameFnt{\small}
\SetAlCapHSkip{0pt}

\acmJournal{TOG}
\begin{document}
% Title portion
% \title{Retrieval-augmented Story-to-Video Generation via Controllable Latent Video Diffusion Models}
%\title{Animate-A-Story: Re-rendering Multi-Source Videos for Visually Coherent Storytelling}
% \title{Animate-A-Story: Exploit Multi-Source Videos for Storytelling Video Synthesis}
\title{Animate-A-Story: \\Storytelling with Retrieval-Augmented Video Generation}

% DO NOT ENTER AUTHOR INFORMATION FOR ANONYMOUS TECHNICAL PAPER SUBMISSIONS TO SIGGRAPH 2019!
\author{Yingqing HE}
\authornote{First authors}
\affiliation{%
 \institution{Hong Kong University of Science and Technology}
 \country{China}}
\author{Menghan Xia}
\authornotemark[1]
\affiliation{%
 \institution{Tencent AI Lab}
 \country{China}
}
\author{Haoxin Chen}
\authornotemark[1]
\affiliation{%
 \institution{Tencent AI Lab}
 \country{China}
}
\author{Xiaodong Cun}
\affiliation{%
 \institution{Tencent AI Lab}
 \country{China}
}
\author{Yuan Gong}
\affiliation{%
 \institution{Tsinghua Shenzhen International Graduate School, Tsinghua University}
 \country{China}
}
\author{Jinbo Xing}
\affiliation{%
 \institution{The Chinese University of Hong Kong}
 \country{China}
}
\author{Yong Zhang}
\authornote{Corresponding authors\\Project page~: ~\url{https://videocrafter.github.io/Animate-A-Story}}
\affiliation{%
 \institution{Tencent AI Lab}
 \country{China}
}
\author{Xintao Wang}
\affiliation{%
 \institution{Tencent AI Lab}
 \country{China}
}
\author{Chao Weng}
\affiliation{%
 \institution{Tencent AI Lab}
 \country{China}
}
\author{Ying Shan}
\affiliation{%
 \institution{Tencent AI Lab}
 \country{China}
}
\author{Qifeng Chen}
\authornotemark[2]
\affiliation{%
 \institution{Hong Kong University of Science and Technology}
 \country{China}
}

\makeatletter
\let\@authorsaddresses\@empty
\makeatother

\begin{abstract}
\if 0
As a favored storytelling medium, movies are widely admired for their immersive and captivating visual experiences. However, movie production is a complex and arduous process that requires professional expertise.
This paper presents an initial exploration of the concept of storyline video synthesis, which involves transforming text descriptions of stories into coherent video content. To achieve this, we propose a controllable video generation system that controls the content synthesis, including structure, 
%style, 
and characters. The system utilizes text prompts and structural guidance to make the generated video in line with the plot, while
% visual style and 
character IDs are further customized to ensure consistency across episodes.
To represent a specific concept, we propose to learn timestep-dependent textural tokens, which greatly improve concept fidelity. Our experiments demonstrate the effectiveness of our proposed methods and show significant performance improvements over existing baselines. In addition, user studies are conducted to evaluate our synthesized storyline videos, which show promising potential for practical applications.
Overall, it suggests that our system could serve as a new tool for creating compelling storytelling videos.
\fi

Generating videos for visual storytelling can be a tedious and complex process that typically requires either live-action filming or graphics animation rendering.
To bypass these challenges, our key idea is to utilize the abundance of existing video clips and synthesize a coherent storytelling video by customizing their appearances. 
We achieve this by developing a framework comprised of two functional modules: (i) Motion Structure Retrieval, which provides video candidates with desired scene or motion context described by query texts, and (ii) Structure-Guided Text-to-Video Synthesis, which generates plot-aligned videos under the guidance of motion structure and text prompts.
For the first module, we leverage an off-the-shelf video retrieval system and extract video depths as motion structure. 
For the second module, we propose a controllable video generation model that offers flexible controls over structure and characters. The videos are synthesized by following the structural guidance and appearance instruction. To ensure visual consistency across clips, we propose an effective concept personalization approach, which allows the specification of the desired character identities through text prompts.
Extensive experiments demonstrate that our approach exhibits significant advantages over various existing baselines.
% Experiments comparing our proposed methods with various existing baselines demonstrate significant advantages of our approach.

% Moreover, user studies on our synthesized storytelling videos demonstrate the effectiveness of our framework and indicate the promising potential for practical applications.
\end{abstract}





\begin{teaserfigure}
  \centering
  \includegraphics[width=1.0\textwidth]{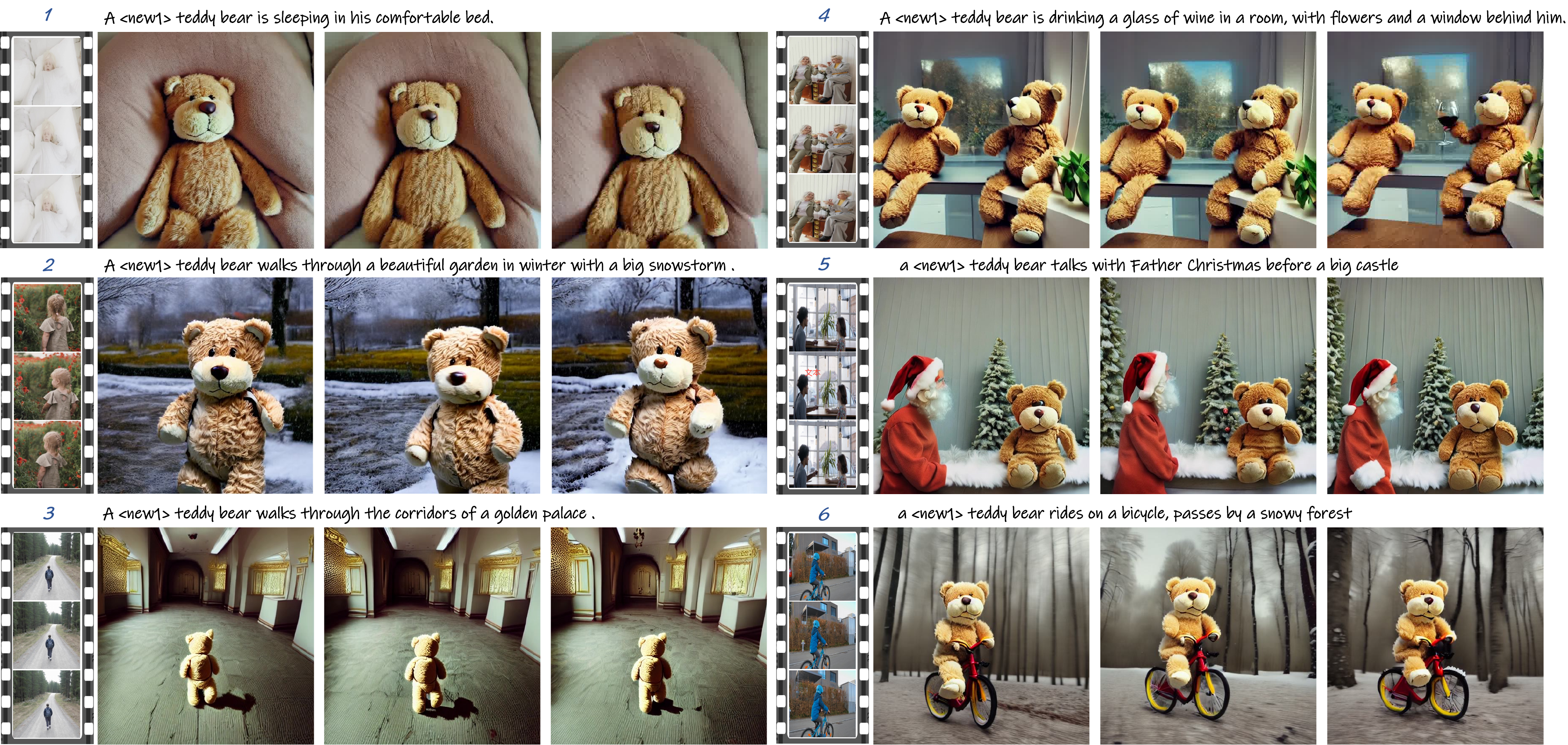}\vspace{-0.5em}
  \caption{Our synthesized videos with a consistent character (user-provided) and storyline. By utilizing text-retrieved video clips as structure guidance, our system manages to generate high-quality videos of similar structures, while also accommodating arbitrary scene appearances based on text prompts. 
  Furthermore, our proposed concept personalization method ensures consistent character rendering across different plot scenarios. 
  Each clip is visualized with three keyframes.
  }
  \label{fig:teaser}
\end{teaserfigure}

\maketitle
\newcommand{\todo}[1]{\textcolor{red}{TODO #1}}
\newcommand{\tocite}{\textcolor{red}{TO Cite}}
\newcommand{\modelname}{LVDM\xspace}

\newcommand{\clip}{c_\theta}
\newcommand{\token}{S_*}
\newcommand{\tokenembd}{v_*}
\newcommand{\tokenembdtable}{V}

\newcommand{\figleft}{{\em (Left)}}
\newcommand{\figcenter}{{\em (Center)}}
\newcommand{\figright}{{\em (Right)}}
\newcommand{\figtop}{{\em (Top)}}
\newcommand{\figbottom}{{\em (Bottom)}}
\newcommand{\captiona}{{\em (a)}}
\newcommand{\captionb}{{\em (b)}}
\newcommand{\captionc}{{\em (c)}}
\newcommand{\captiond}{{\em (d)}}

\newcommand{\newterm}[1]{{\bf #1}}

\def\figref#1{figure~\ref{#1}}
\def\Figref#1{Figure~\ref{#1}}
\def\twofigref#1#2{figures \ref{#1} and \ref{#2}}
\def\quadfigref#1#2#3#4{figures \ref{#1}, \ref{#2}, \ref{#3} and \ref{#4}}
\def\secref#1{section~\ref{#1}}
\def\Secref#1{Section~\ref{#1}}
\def\twosecrefs#1#2{sections \ref{#1} and \ref{#2}}
\def\secrefs#1#2#3{sections \ref{#1}, \ref{#2} and \ref{#3}}
\def\eqref#1{equation~\ref{#1}}
\def\Eqref#1{Equation~\ref{#1}}
\def\plaineqref#1{\ref{#1}}
\def\chapref#1{chapter~\ref{#1}}
\def\Chapref#1{Chapter~\ref{#1}}
\def\rangechapref#1#2{chapters\ref{#1}--\ref{#2}}
\def\algref#1{algorithm~\ref{#1}}
\def\Algref#1{Algorithm~\ref{#1}}
\def\twoalgref#1#2{algorithms \ref{#1} and \ref{#2}}
\def\Twoalgref#1#2{Algorithms \ref{#1} and \ref{#2}}
\def\partref#1{part~\ref{#1}}
\def\Partref#1{Part~\ref{#1}}
\def\twopartref#1#2{parts \ref{#1} and \ref{#2}}

\def\ceil#1{\lceil #1 \rceil}
\def\floor#1{\lfloor #1 \rfloor}
\def\1{\bm{1}}
\newcommand{\train}{\mathcal{D}}
\newcommand{\valid}{\mathcal{D_{\mathrm{valid}}}}
\newcommand{\test}{\mathcal{D_{\mathrm{test}}}}

\def\eps{{\epsilon}}

\def\reta{{\textnormal{$\eta$}}}
\def\ra{{\textnormal{a}}}
\def\rb{{\textnormal{b}}}
\def\rc{{\textnormal{c}}}
\def\rd{{\textnormal{d}}}
\def\re{{\textnormal{e}}}
\def\rf{{\textnormal{f}}}
\def\rg{{\textnormal{g}}}
\def\rh{{\textnormal{h}}}
\def\ri{{\textnormal{i}}}
\def\rj{{\textnormal{j}}}
\def\rk{{\textnormal{k}}}
\def\rl{{\textnormal{l}}}
\def\rn{{\textnormal{n}}}
\def\ro{{\textnormal{o}}}
\def\rp{{\textnormal{p}}}
\def\rq{{\textnormal{q}}}
\def\rr{{\textnormal{r}}}
\def\rs{{\textnormal{s}}}
\def\rt{{\textnormal{t}}}
\def\ru{{\textnormal{u}}}
\def\rv{{\textnormal{v}}}
\def\rw{{\textnormal{w}}}
\def\rx{{\textnormal{x}}}
\def\ry{{\textnormal{y}}}
\def\rz{{\textnormal{z}}}

\def\rvepsilon{{\mathbf{\epsilon}}}
\def\rvtheta{{\mathbf{\theta}}}
\def\rva{{\mathbf{a}}}
\def\rvb{{\mathbf{b}}}
\def\rvc{{\mathbf{c}}}
\def\rvd{{\mathbf{d}}}
\def\rve{{\mathbf{e}}}
\def\rvf{{\mathbf{f}}}
\def\rvg{{\mathbf{g}}}
\def\rvh{{\mathbf{h}}}
\def\rvu{{\mathbf{i}}}
\def\rvj{{\mathbf{j}}}
\def\rvk{{\mathbf{k}}}
\def\rvl{{\mathbf{l}}}
\def\rvm{{\mathbf{m}}}
\def\rvn{{\mathbf{n}}}
\def\rvo{{\mathbf{o}}}
\def\rvp{{\mathbf{p}}}
\def\rvq{{\mathbf{q}}}
\def\rvr{{\mathbf{r}}}
\def\rvs{{\mathbf{s}}}
\def\rvt{{\mathbf{t}}}
\def\rvu{{\mathbf{u}}}
\def\rvv{{\mathbf{v}}}
\def\rvw{{\mathbf{w}}}
\def\rvx{{\mathbf{x}}}
\def\rvy{{\mathbf{y}}}
\def\rvz{{\mathbf{z}}}

\def\erva{{\textnormal{a}}}
\def\ervb{{\textnormal{b}}}
\def\ervc{{\textnormal{c}}}
\def\ervd{{\textnormal{d}}}
\def\erve{{\textnormal{e}}}
\def\ervf{{\textnormal{f}}}
\def\ervg{{\textnormal{g}}}
\def\ervh{{\textnormal{h}}}
\def\ervi{{\textnormal{i}}}
\def\ervj{{\textnormal{j}}}
\def\ervk{{\textnormal{k}}}
\def\ervl{{\textnormal{l}}}
\def\ervm{{\textnormal{m}}}
\def\ervn{{\textnormal{n}}}
\def\ervo{{\textnormal{o}}}
\def\ervp{{\textnormal{p}}}
\def\ervq{{\textnormal{q}}}
\def\ervr{{\textnormal{r}}}
\def\ervs{{\textnormal{s}}}
\def\ervt{{\textnormal{t}}}
\def\ervu{{\textnormal{u}}}
\def\ervv{{\textnormal{v}}}
\def\ervw{{\textnormal{w}}}
\def\ervx{{\textnormal{x}}}
\def\ervy{{\textnormal{y}}}
\def\ervz{{\textnormal{z}}}

\def\rmA{{\mathbf{A}}}
\def\rmB{{\mathbf{B}}}
\def\rmC{{\mathbf{C}}}
\def\rmD{{\mathbf{D}}}
\def\rmE{{\mathbf{E}}}
\def\rmF{{\mathbf{F}}}
\def\rmG{{\mathbf{G}}}
\def\rmH{{\mathbf{H}}}
\def\rmI{{\mathbf{I}}}
\def\rmJ{{\mathbf{J}}}
\def\rmK{{\mathbf{K}}}
\def\rmL{{\mathbf{L}}}
\def\rmM{{\mathbf{M}}}
\def\rmN{{\mathbf{N}}}
\def\rmO{{\mathbf{O}}}
\def\rmP{{\mathbf{P}}}
\def\rmQ{{\mathbf{Q}}}
\def\rmR{{\mathbf{R}}}
\def\rmS{{\mathbf{S}}}
\def\rmT{{\mathbf{T}}}
\def\rmU{{\mathbf{U}}}
\def\rmV{{\mathbf{V}}}
\def\rmW{{\mathbf{W}}}
\def\rmX{{\mathbf{X}}}
\def\rmY{{\mathbf{Y}}}
\def\rmZ{{\mathbf{Z}}}

\def\ermA{{\textnormal{A}}}
\def\ermB{{\textnormal{B}}}
\def\ermC{{\textnormal{C}}}
\def\ermD{{\textnormal{D}}}
\def\ermE{{\textnormal{E}}}
\def\ermF{{\textnormal{F}}}
\def\ermG{{\textnormal{G}}}
\def\ermH{{\textnormal{H}}}
\def\ermI{{\textnormal{I}}}
\def\ermJ{{\textnormal{J}}}
\def\ermK{{\textnormal{K}}}
\def\ermL{{\textnormal{L}}}
\def\ermM{{\textnormal{M}}}
\def\ermN{{\textnormal{N}}}
\def\ermO{{\textnormal{O}}}
\def\ermP{{\textnormal{P}}}
\def\ermQ{{\textnormal{Q}}}
\def\ermR{{\textnormal{R}}}
\def\ermS{{\textnormal{S}}}
\def\ermT{{\textnormal{T}}}
\def\ermU{{\textnormal{U}}}
\def\ermV{{\textnormal{V}}}
\def\ermW{{\textnormal{W}}}
\def\ermX{{\textnormal{X}}}
\def\ermY{{\textnormal{Y}}}
\def\ermZ{{\textnormal{Z}}}

\def\vzero{{\bm{0}}}
\def\vone{{\bm{1}}}
\def\vmu{{\bm{\mu}}}
\def\vtheta{{\bm{\theta}}}
\def\va{{\bm{a}}}
\def\vb{{\bm{b}}}
\def\vc{{\bm{c}}}
\def\vd{{\bm{d}}}
\def\ve{{\bm{e}}}
\def\vf{{\bm{f}}}
\def\vg{{\bm{g}}}
\def\vh{{\bm{h}}}
\def\vi{{\bm{i}}}
\def\vj{{\bm{j}}}
\def\vk{{\bm{k}}}
\def\vl{{\bm{l}}}
\def\vm{{\bm{m}}}
\def\vn{{\bm{n}}}
\def\vo{{\bm{o}}}
\def\vp{{\bm{p}}}
\def\vq{{\bm{q}}}
\def\vr{{\bm{r}}}
\def\vs{{\bm{s}}}
\def\vt{{\bm{t}}}
\def\vu{{\bm{u}}}
\def\vv{{\bm{v}}}
\def\vw{{\bm{w}}}
\def\vx{{\bm{x}}}
\def\vy{{\bm{y}}}
\def\vz{{\bm{z}}}

\def\evalpha{{\alpha}}
\def\evbeta{{\beta}}
\def\evepsilon{{\epsilon}}
\def\evlambda{{\lambda}}
\def\evomega{{\omega}}
\def\evmu{{\mu}}
\def\evpsi{{\psi}}
\def\evsigma{{\sigma}}
\def\evtheta{{\theta}}
\def\eva{{a}}
\def\evb{{b}}
\def\evc{{c}}
\def\evd{{d}}
\def\eve{{e}}
\def\evf{{f}}
\def\evg{{g}}
\def\evh{{h}}
\def\evi{{i}}
\def\evj{{j}}
\def\evk{{k}}
\def\evl{{l}}
\def\evm{{m}}
\def\evn{{n}}
\def\evo{{o}}
\def\evp{{p}}
\def\evq{{q}}
\def\evr{{r}}
\def\evs{{s}}
\def\evt{{t}}
\def\evu{{u}}
\def\evv{{v}}
\def\evw{{w}}
\def\evx{{x}}
\def\evy{{y}}
\def\evz{{z}}

\def\mA{{\bm{A}}}
\def\mB{{\bm{B}}}
\def\mC{{\bm{C}}}
\def\mD{{\bm{D}}}
\def\mE{{\bm{E}}}
\def\mF{{\bm{F}}}
\def\mG{{\bm{G}}}
\def\mH{{\bm{H}}}
\def\mI{{\bm{I}}}
\def\mJ{{\bm{J}}}
\def\mK{{\bm{K}}}
\def\mL{{\bm{L}}}
\def\mM{{\bm{M}}}
\def\mN{{\bm{N}}}
\def\mO{{\bm{O}}}
\def\mP{{\bm{P}}}
\def\mQ{{\bm{Q}}}
\def\mR{{\bm{R}}}
\def\mS{{\bm{S}}}
\def\mT{{\bm{T}}}
\def\mU{{\bm{U}}}
\def\mV{{\bm{V}}}
\def\mW{{\bm{W}}}
\def\mX{{\bm{X}}}
\def\mY{{\bm{Y}}}
\def\mZ{{\bm{Z}}}
\def\mBeta{{\bm{\beta}}}
\def\mPhi{{\bm{\Phi}}}
\def\mLambda{{\bm{\Lambda}}}
\def\mSigma{{\bm{\Sigma}}}

% \DeclareMathAlphabet{\mathsfit}{\encodingdefault}{\sfdefault}{m}{sl}
% \SetMathAlphabet{\mathsfit}{bold}{\encodingdefault}{\sfdefault}{bx}{n}
\newcommand{\tens}[1]{\bm{\mathsfit{#1}}}
\def\tA{{\tens{A}}}
\def\tB{{\tens{B}}}
\def\tC{{\tens{C}}}
\def\tD{{\tens{D}}}
\def\tE{{\tens{E}}}
\def\tF{{\tens{F}}}
\def\tG{{\tens{G}}}
\def\tH{{\tens{H}}}
\def\tI{{\tens{I}}}
\def\tJ{{\tens{J}}}
\def\tK{{\tens{K}}}
\def\tL{{\tens{L}}}
\def\tM{{\tens{M}}}
\def\tN{{\tens{N}}}
\def\tO{{\tens{O}}}
\def\tP{{\tens{P}}}
\def\tQ{{\tens{Q}}}
\def\tR{{\tens{R}}}
\def\tS{{\tens{S}}}
\def\tT{{\tens{T}}}
\def\tU{{\tens{U}}}
\def\tV{{\tens{V}}}
\def\tW{{\tens{W}}}
\def\tX{{\tens{X}}}
\def\tY{{\tens{Y}}}
\def\tZ{{\tens{Z}}}

\def\gA{{\mathcal{A}}}
\def\gB{{\mathcal{B}}}
\def\gC{{\mathcal{C}}}
\def\gD{{\mathcal{D}}}
\def\gE{{\mathcal{E}}}
\def\gF{{\mathcal{F}}}
\def\gG{{\mathcal{G}}}
\def\gH{{\mathcal{H}}}
\def\gI{{\mathcal{I}}}
\def\gJ{{\mathcal{J}}}
\def\gK{{\mathcal{K}}}
\def\gL{{\mathcal{L}}}
\def\gM{{\mathcal{M}}}
\def\gN{{\mathcal{N}}}
\def\gO{{\mathcal{O}}}
\def\gP{{\mathcal{P}}}
\def\gQ{{\mathcal{Q}}}
\def\gR{{\mathcal{R}}}
\def\gS{{\mathcal{S}}}
\def\gT{{\mathcal{T}}}
\def\gU{{\mathcal{U}}}
\def\gV{{\mathcal{V}}}
\def\gW{{\mathcal{W}}}
\def\gX{{\mathcal{X}}}
\def\gY{{\mathcal{Y}}}
\def\gZ{{\mathcal{Z}}}

\def\sA{{\mathbb{A}}}
\def\sB{{\mathbb{B}}}
\def\sC{{\mathbb{C}}}
\def\sD{{\mathbb{D}}}
\def\sF{{\mathbb{F}}}
\def\sG{{\mathbb{G}}}
\def\sH{{\mathbb{H}}}
\def\sI{{\mathbb{I}}}
\def\sJ{{\mathbb{J}}}
\def\sK{{\mathbb{K}}}
\def\sL{{\mathbb{L}}}
\def\sM{{\mathbb{M}}}
\def\sN{{\mathbb{N}}}
\def\sO{{\mathbb{O}}}
\def\sP{{\mathbb{P}}}
\def\sQ{{\mathbb{Q}}}
\def\sR{{\mathbb{R}}}
\def\sS{{\mathbb{S}}}
\def\sT{{\mathbb{T}}}
\def\sU{{\mathbb{U}}}
\def\sV{{\mathbb{V}}}
\def\sW{{\mathbb{W}}}
\def\sX{{\mathbb{X}}}
\def\sY{{\mathbb{Y}}}
\def\sZ{{\mathbb{Z}}}

\def\emLambda{{\Lambda}}
\def\emA{{A}}
\def\emB{{B}}
\def\emC{{C}}
\def\emD{{D}}
\def\emE{{E}}
\def\emF{{F}}
\def\emG{{G}}
\def\emH{{H}}
\def\emI{{I}}
\def\emJ{{J}}
\def\emK{{K}}
\def\emL{{L}}
\def\emM{{M}}
\def\emN{{N}}
\def\emO{{O}}
\def\emP{{P}}
\def\emQ{{Q}}
\def\emR{{R}}
\def\emS{{S}}
\def\emT{{T}}
\def\emU{{U}}
\def\emV{{V}}
\def\emW{{W}}
\def\emX{{X}}
\def\emY{{Y}}
\def\emZ{{Z}}
\def\emSigma{{\Sigma}}

\newcommand{\etens}[1]{\mathsfit{#1}}
\def\etLambda{{\etens{\Lambda}}}
\def\etA{{\etens{A}}}
\def\etB{{\etens{B}}}
\def\etC{{\etens{C}}}
\def\etD{{\etens{D}}}
\def\etE{{\etens{E}}}
\def\etF{{\etens{F}}}
\def\etG{{\etens{G}}}
\def\etH{{\etens{H}}}
\def\etI{{\etens{I}}}
\def\etJ{{\etens{J}}}
\def\etK{{\etens{K}}}
\def\etL{{\etens{L}}}
\def\etM{{\etens{M}}}
\def\etN{{\etens{N}}}
\def\etO{{\etens{O}}}
\def\etP{{\etens{P}}}
\def\etQ{{\etens{Q}}}
\def\etR{{\etens{R}}}
\def\etS{{\etens{S}}}
\def\etT{{\etens{T}}}
\def\etU{{\etens{U}}}
\def\etV{{\etens{V}}}
\def\etW{{\etens{W}}}
\def\etX{{\etens{X}}}
\def\etY{{\etens{Y}}}
\def\etZ{{\etens{Z}}}

\newcommand{\pdata}{p_{\rm{data}}}
\newcommand{\ptrain}{\hat{p}_{\rm{data}}}
\newcommand{\Ptrain}{\hat{P}_{\rm{data}}}
\newcommand{\pmodel}{p_{\rm{model}}}
\newcommand{\Pmodel}{P_{\rm{model}}}
\newcommand{\ptildemodel}{\tilde{p}_{\rm{model}}}
\newcommand{\pencode}{p_{\rm{encoder}}}
\newcommand{\pdecode}{p_{\rm{decoder}}}
\newcommand{\precons}{p_{\rm{reconstruct}}}

\newcommand{\laplace}{\mathrm{Laplace}} %

\newcommand{\E}{\mathbb{E}}
\newcommand{\Ls}{\mathcal{L}}
\newcommand{\R}{\mathbb{R}}
\newcommand{\emp}{\tilde{p}}
\newcommand{\lr}{\alpha}
\newcommand{\reg}{\lambda}
\newcommand{\rect}{\mathrm{rectifier}}
\newcommand{\softmax}{\mathrm{softmax}}
\newcommand{\sigmoid}{\sigma}
\newcommand{\softplus}{\zeta}
\newcommand{\KL}{D_{\mathrm{KL}}}
\newcommand{\Var}{\mathrm{Var}}
\newcommand{\standarderror}{\mathrm{SE}}
\newcommand{\Cov}{\mathrm{Cov}}
\newcommand{\normlzero}{L^0}
\newcommand{\normlone}{L^1}
\newcommand{\normltwo}{L^2}
\newcommand{\normlp}{L^p}
\newcommand{\normmax}{L^\infty}

\newcommand{\parents}{Pa} %

\let\ab\allowbreak

%---------------------------------------------
\section{Introduction}
\label{sec:intro}
%---------------------------------------------

%It is a tedious process to produce a film according to the  story scripts, which usually involves xxx. Recently the progress made in generative fundamental models pushes forward the performance of video generation largely. With text2video model, we can obtain visual plausible video clips by typing text prompts, which intrigues us to explore the possibility of producing films by video synthesis purely.

%Current text2video models show remarkable results. However, in most cases, text prompt is not enough to guide the model to synthesize a video that fulfills our customized needs, such as presenting a film scene. Specifically, we need more controls, such as structure, style, and characters, over the content synthesis, so as to make video clips that are qualified to compose a film with content consistency.
Creating engaging storytelling videos is a complex and laborious process that typically involves live-action filming or CG animation production. 
This technical nature not only demands significant resources from professional content creators but also creates barriers for the general public in effectively utilizing this powerful medium. 
Recently, significant progress has been made in text-to-video (T2V) generation, allowing for the automatic generation of videos based on textual descriptions~\cite{ho2022imagen, zhou2022magicvideo, he2022latent, singer2022make}. 

However, the effectiveness of these video generation techniques is still limited, yielding results that fall short of expectations and hinder their practical application. 
Additionally, the layout and composition of the generated video cannot be controlled through text, which is crucial for visualizing an appealing story and filming a movie. For example, close-ups, long shots, and composition can assist directors in conveying implicit information to the audience.
Current text-to-video generation models can hardly generate proper motions and layouts that meet the requirement of film. 

To overcome these challenges, we propose a novel video generation approach incorporating the abundance of existing video content into the T2V generation process, which we refer to as \textit{retrieval-augmented video generation}.
Specifically, our approach retrieves videos from external databases based on text prompts and utilizes them as a guidance signal for the T2V generation.
Building on this idea, our approach also enables users to have greater control over the layout and composition of the generated videos when animating a story, by utilizing the input retrieved videos as a structure reference. 
Additionally, the quality of the generated videos is enhanced by leveraging the rich knowledge and information contained in the retrieved videos, which can be used to improve the realism and coherence of the generated scenes.
% Furthermore, our approach allows for the re-rendering of the appearances of scenes and objects based on the given text prompt, enabling the generation of novel videos that are tailored to the specific requirements of the user. 
The text prompt is now responsible for rendering the appearances of scenes and objects to generate novel videos.
% Specifically, we retrieve video clips featuring desired scenario contexts through query texts and customize their appearances to construct a visually coherent storyline video based on storyboard scripts.
% To facilitate this, we develop a storyline video synthesis framework consisting of three functional modules: motion structure retrieval, structure-guided text-to-video synthesis, and personalized character rerendering.
% First, we utilize an off-the-shelf video retrieval engine to gather video clips based on text prompts to display the desired actions and composition of a picture required by the story. 

% By incorporating these features, our approach provides a more efficient and accessible way for content creators to produce high-quality animated videos, while also reducing the time and resources required for video production.

However, such retrieval-guided video generation processes still suffer from the inconsistency problem of the character across different video clips.
Besides, the character's appearance is controlled by the text prompt and generated in a stochastic way which lacks user control.
To further tackle this issue, we study existing literature on personalization~\cite{ruiz2023dreambooth} to finetune the generation model to re-render the appearance of the character.
and propose a novel approach (TimeInv) to better represent personalized concepts and improve performance.

By incorporating the two core modules: retrieval-enhanced T2V generation and video character rerendering, our approach provides a more efficient and accessible way for content creators to produce high-quality animated videos. To justify its effectiveness, we evaluate our method from the following perspectives:
First, our retrieval-enhanced T2V generation model is compared against existing baselines, demonstrating notable superiority in video generation performance. 
Second, we justify the advantage of our proposed personalization method in relation to existing competitors. 
Furthermore, we conduct comprehensive experiments of the overall effectiveness of our proposed storytelling video synthesis framework, suggesting its potential for practical applications.

Our contribution is summarized as the following:
\begin{itemize}
    \item We present a novel retrieval-augmented paradigm for storytelling video synthesis, which enables the use of existing diverse videos for storytelling purposes for the first time. 
    Experimental results demonstrate the framework's effectiveness, positioning it as a novel video-making tool with remarkable convenience.
    \item We propose an adjustable structure-guided text-to-video model, which effectively resolves the conflict between structure guidance and character generation.
    \item We propose TimeInv, a new concept personalization approach that outperforms existing competitors notably. 
\end{itemize}
% style, 
%---------------------------------------------
\section{Related Work}
\label{sec:intro}
%---------------------------------------------
\subsection{Video Generation}

Numerous earlier works~\cite{vondrick2016generating,saito2017temporal,tulyakov2018mocogan,wu2021godiva,skorokhodov2022stylegan} focus on unconditional video generation methods, employing generative adversarial networks (GANs) or variational auto-encoder (VAE) to model video distributions. For instance, VGAN~\cite{vondrick2016generating} separately models the foreground and background by mapping random noise to space-time cuboid feature maps. The fused feature map signifies the generated video and is input into a discriminator. TGAN~\cite{saito2017temporal} generates a noise sequence with a temporal generator and then uses an image generator to transform it into images. These images are concatenated and fed into a video discriminator. StyleGAN-V~\cite{skorokhodov2022stylegan} leverages the capabilities of StyleGAN. It uses multiple random noises to control motion and an additional noise to manage appearance.

Several methods~\cite{yan2021videogpt,ge2022long,yu2023magvit} aim to capture spatio-temporal dependencies using transformers in the latent space. Initially, they project videos into a latent space by learning a VAE or VQGAN~\cite{esser2021taming}, followed by training a transformer to model the latent distribution. For example, TATS~\cite{ge2022long} trains a time-agnostic VQGAN and subsequently learns a time-sensitive transformer based on the latent features. VideoGPT~\cite{yan2021videogpt} follows a similar pipeline. To enable text control over generated content, some works~\cite{hong2022cogvideo,villegas2022phenaki} extract visual and text tokens, projecting them into the same latent space. A transformer is consistently used to model the interdependencies among them.

% Recently, text-to-image (T2I) generation~\cite{rombach2022high,saharia2022photorealistic,ramesh2022hierarchical} has accomplished great advances in generating high-quality images, especially diffusion-based models. 
% Benefiting from the evolution of T2I generation, the field of text-to-video generation have also achieve breakthroughs. 
% VDM~\cite{ho2022video} is the first work that uses diffusion models for video generation. 
% Make-a-video~\cite{singer2022make} and Imagen Video~\cite{ho2022imagen} are cascade models that model video distribution at low-resolution first and than apply spatio-temporal interpolation to enlarge the resolution and time duration. 
% Inspired by LDM~\cite{rombach2022high}, a set of works~\cite{he2022latent,zhou2022magicvideo,mei2022vidm,yu2023video,luo2023decomposed,blattmann2023align} extent LDM for video generation. 
% For instance, LVDM~\cite{he2022latent} inflates LDM into a video version by introducing temporal attention layers. It uses the pretrained LDM as initialization and trains the learnable parameters with video data. 
% Similar to LDM, the text embedding is injected into the UNet by using the cross-attention mechanism. 
% Video LDM~\cite{blattmann2023align} shares the same idea with LVDM, but with
% the difference that it fixes the spatial weights of LDM. 

Recently, text-to-image (T2I) generation~\cite{rombach2022high,saharia2022photorealistic,ramesh2022hierarchical} has achieved significant advancements in generating high-quality images, primarily due to diffusion-based models. Capitalizing on the progress of T2I generation, the field of text-to-video generation has experienced breakthroughs as well. VDM~\cite{ho2022video} is the first work employing diffusion models for video generation. Make-a-video~\cite{singer2022make} and Imagen Video~\cite{ho2022imagen} are cascade models that initially model video distribution at low-resolution and then apply spatio-temporal interpolation to increase the resolution and time duration. Inspired by LDM~\cite{rombach2022high}, several works~\cite{he2022latent,zhou2022magicvideo,mei2022vidm,yu2023video,luo2023decomposed,blattmann2023align} extend LDM for video generation. For example, LVDM~\cite{he2022latent} inflates LDM into a video version by introducing temporal attention layers. It employs the pretrained LDM as initialization and trains the learnable parameters with video data. Similar to LDM, the text embedding is injected into the UNet using the cross-attention mechanism. Video LDM~\cite{blattmann2023align} shares the same concept as LVDM, but with the distinction that it fixes the spatial weights of LDM.

\subsection{Structure-guided Video Generation}
% Similar to the evolution of T2I, a lot of works~\cite{xing2023make,yang2023rerender,wang2023genlvideo,zhang2023controlvideo,wang2023videocomposer} explore the capability of conditional video generation based on pretrained text-to-image or text-to-video models.  
% For example, Make-Your-Video~\cite{xing2023make} uses depth as the additional condition besides the text. The spatial weights of Stable Diffusion is fixed while the newly added temporal weights are learned on video data. 
% As depth is extracted from video, it can rerender the appearance of the source video. 
% VideoComposer~\cite{wang2023videocomposer} is an extension of Composer~\cite{huang2023composer} that takes multiple types of images as condition such as RGB image, sketch, depth, etc.. 
% Those conditions are fused in the latent space and interact with the UNet via cross attention. 
Mirroring the evolution of T2I, numerous works~\cite{xing2023make,yang2023rerender,wang2023genlvideo,zhang2023controlvideo,wang2023videocomposer} investigate the capability of conditional video generation based on pretrained text-to-image or text-to-video models. For instance, Make-Your-Video~\cite{xing2023make} uses depth as an additional condition besides text. The spatial weights of Stable Diffusion are fixed, while the newly added temporal weights are learned on video data. Since depth is extracted from the video, it can re-render the appearance of the source video. 
Follow-Your-Pose~\cite{ma2023followyourpose} utilize pose as a condition to guide the human-like character video synthesis process.
VideoComposer~\cite{wang2023videocomposer}, an extension of Composer~\cite{huang2023composer}, takes multiple types of images as conditions, such as RGB images, sketches, depths, etc. These conditions are fused in the latent space and interact with the UNet via cross attention.

\subsection{Concept Customization}
% Generating an image with the specified object is called customization or personalization. 
% Many works~\cite{textualinv,ruiz2023dreambooth,customdiffusion,wei2023elite,alaluf2023neural} focus on this task from different perspectives. 
% Textual inversion~\cite{textualinv} is the first inversion work on Stable Diffusion. It optimizes a token without tuning the model for the given images of an object. 
% Dreambooth~\cite{textualinv} not only learns a token but also finetunes the whole model parameters.  
% Multi-concept~\cite{customdiffusion} is the first to propose a method for the inversion of multiple concepts. 
% ELITE~\cite{wei2023elite} trains an encoder to map visual image to text embedding for customized text-to-image generation, instead of optimization. 
% NeTI~\cite{alaluf2023neural} introduces a new text-conditioning latent space that is dependent on both timestep and UNet layers. 

Generating an image with a specified object is referred to as customization or personalization. Numerous works~\cite{textualinv,ruiz2023dreambooth,customdiffusion,wei2023elite,alaluf2023neural} explore this task from various perspectives. Textual inversion~\cite{textualinv}, the first inversion work on Stable Diffusion, optimizes a token without tuning the model for a given object's images. In contrast, Dreambooth~\cite{textualinv} learns a token and fine-tunes the entire model parameters. Multi-concept~\cite{customdiffusion} is the first to propose a method for inverting multiple concepts. ELITE~\cite{wei2023elite} trains an encoder to map visual images to text embeddings for customized text-to-image generation, rather than optimization. NeTI~\cite{alaluf2023neural} introduces a novel text-conditioning latent space that depends on both the timestep and UNet layers. It learns a mapping that projects timestep and layer index into the embedding space. 
% \subsection{Story-to-video generation}
% 1.text2video generation 

% 2.structure-controllable video generation 
% % gen1, videocrafter-adapter, text2video zero

% 3.Concept Customization

%---------------------------------------------
\section{Method}
\label{subsec:method}
%---------------------------------------------

Our goal is to develop a framework that can automatically generate high-quality storytelling videos based on storyline scripts or with minimal interactive effort. 
To achieve this, we propose to retrieve existing video assets to enhance the performance of T2V generation (see \cref{subsec:overview}).
Specifically, we extract the structures from retrieved videos, which will then serve as guidance signals provided to the T2V process (see \cref{subsec:controllable_synthesis}).
Additionally, we perform video character rerendering based on the proposed TimeInv approach to synthesize consistent characters across different video clips (see \cref{subsec:coherent_synthesis}).
In the following sections, we will delve into the key technical designs that enable its functionality.

\begin{figure}[!t]
	\centering
	\includegraphics[width=1\linewidth]{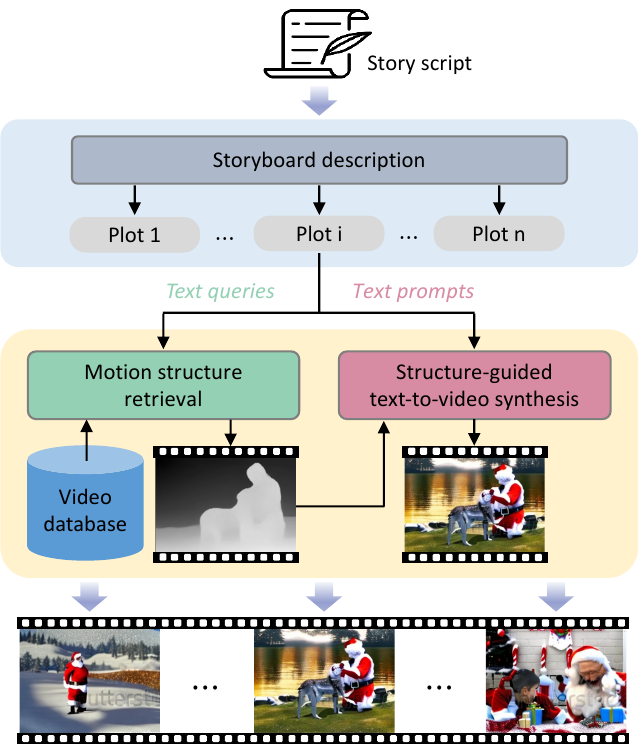}\vspace{-0.5em}
	\caption{Flowchart of our retrieval-augmented video synthesis framework. 
    Given a textual story script, we first extract the key plots and modulate their descriptions as text queries and prompts. 
    Each plot is transformed into a generated video clip through two modules: a video retrieval system and a structure-guided text-to-video model. 
    % Lastly, the synthesized video clips are arranged sequentially to present a comprehensive storyline video that vividly conveys the story's narrative.
    }
    \label{fig:system_flowchart}
\end{figure}

%---------------------------------------------
\subsection{Retrieval-augmented Text-to-Video Generation}
\label{subsec:overview}
%---------------------------------------------

As illustrated in Fig.~\ref{fig:system_flowchart}, our video generation framework involves three procedures: text processing, video retrieval, and video synthesis. 
In the text processing stage, we extract the key plots from the story script through storyboard analysis. To simplify the problem, we regulate an individual plot as a single event without a shot transition. For instance, "a boy ran into a wolf in the forest" is a single plot, while "a boy ran into a wolf in the forest and he killed the wolf with a gun" should be separated into two plots. 
For each plot, we further adjust and decorate the description so that they can serve as effective text queries and text prompts respectively. 
This stage is completed manually or using the assistance of large language models (LLMs) like GPT-4~\cite{openai2023gpt4}.

\begin{figure}[!t]
	\centering
	\includegraphics[width=1\linewidth]{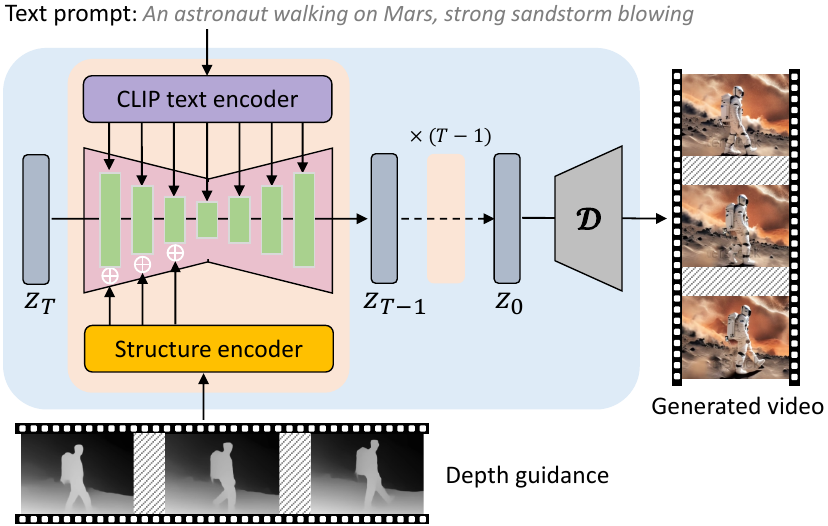}\vspace{-0.5em}
	\caption{Overview of our adjustable structure-guided text-to-video model. 
    We use the depth information from source videos to guide the video synthesis process.
    The model consists of two branches: a general text-to-video synthesis branch which is a video diffusion model in the latent space, and a side branch for encoding and imposing structure control.
    The controlling mechanism is elementwise feature addition.
    Notably, the depth control is \textit{adjustable} and this property is crucial for further character rerendering, which we will illustrate in \cref{subsec:coherent_synthesis}.
    % As a latent diffusion model, the forward diffusion process and reverse denosing process is formulated in the latent space that is constructed by an independently learned VAE. 
    % To allow control over the video synthesis, the denoising process is modelled by conditioning on extra input, i.e. text prompt and frame-wise depth. 
    % In correspondence to the $T$-step diffusion, the sampling of new videos involves an iterative denoising process, starting from a random noise tensor.
    }
    \label{fig:model_overview}
\end{figure}

After that, we process each plot separately using two sequentially conducted modules. 
Given the text query, we can obtain video candidates showing desired scenario through an off-the-shelf text-based video retrieval engine~\cite{brain2021frozen} that associates with a database with about 10M open-world videos collected from the Internet. 
Since the video appearance may not match the plot precisely, we only take the motion structure of it by applying a depth estimation algorithm to it. This extends the usability of existing videos. As the example illustrated in \cref{fig:model_overview}, to synthesize the video of "Santa Claus playing with a \textit{wolf} in the forest", we can use the motion structure in the video of "a man playing with a \textit{dog} in the park", which is quite common in the video database. 
Utilizing the motion structure as guidance, we can synthesize plot-aligned videos through text prompts. Next, we describe the structure-guided T2V model in detail. 
% Notably, customized characters are supported through the proposed TimeInv which will be described in \cref{sec:coherent_synthesis}.
% Finally, all the plot videos are composed as a complete storytelling video with visual coherence.

%---------------------------------------------
\subsection{Structure-Guided Text-to-Video Synthesis}
\label{subsec:controllable_synthesis}
%---------------------------------------------

\textbf{Preliminary.}
Denoising Diffusion Probabilistic Models (DDPM)~\cite{ho2020denoising}, also called Diffusion Models (DM) for short, learn to model an empirical data distribution $p_{data}(\mathbf{x})$ by building the mapping from a standard Gaussian distribution to the target distribution. Particularly, the forward process is formulated as a fixed diffusion process that is denoted as:
\begin{equation}
q(\mathbf{x}_t|\mathbf{x}_{t-1})\coloneqq \mathcal{N}(\mathbf{x}_t, \sqrt{1-\beta_t} \mathbf{x}_{t-1}, \beta_tI).
\end{equation}
This is a Markov chain that adds noise into each sample $\mathbf{x}_0$ gradually with the variance schedule $\beta_t \in (0,1)$ where $t \in \{1, ..., T\}$ with $T$ being the total steps of the diffusion chain.
The reverse process is realized through a denoiser model $\mathbf{f}_{\theta}$ that takes the diffused $\mathbf{x}_t$ together with the current time step $t$ as input and it is optimized by minimizing the denoising score matching objective:
\begin{equation}
L(\theta)=\mathbb{E}_{\mathbf{x}_0\sim p_{data}, t} \lVert \mathbf{\epsilon}_t-\mathbf{f}_{\theta}(\mathbf{x}_t; \mathbf{c}, t) \rVert_2^2,
\end{equation}
where $\mathbf{c}$ is optional conditioning information (e.g. text embedding) and the supervision $\mathbf{\epsilon}_t$ is a randomly sampled noise used in the forward diffusion process: $\mathbf{x}_0 \rightarrow \mathbf{x}_t$.
Extended from DM, LDM formulates the diffusion and denoising process at a latent space. That is usually a compressive representation space that is separated and learned through a Variational Autoencoder(VAE)~\cite{kingma2014auto} comprised of an encoder $\mathcal{E}$ and a decoder $\mathcal{D}$.

We employ a conditional LDM to learn controllable video synthesis, as the overview depicted in Fig.~\ref{fig:model_overview}.
In our approach, the videos are transformed (or reconstructed) into (or from) the latent space in a frame-wise manner. 
Specifically, our encoder $\mathcal{E}$ downsamples the input RGB-image $\mathbf{x} \in \mathbb{R}^{3\times H \times W}$ with a downsampling factor of 8 and outputs a latent representation $\mathbf{z} \in \mathbb{R}^{4 \times H' \times W'}$, where $H' = H /8, W' = W/8$, enabling the denoiser model to work on a much lower dimensional data and thus improves the running time and memory efficiency.
The latent diffusion model is conditioned on both the motion structure and text prompt. Given a video clip $\mathbf{x} \in \mathbb{R}^{L\times 3\times H\times W}$ with $L$ frames, we obtain its latent representation $\mathbf{z}$ through $\mathbf{z} = \mathcal{E}(\mathbf{x})$, where $\mathbf{z} \in \mathbb{R}^{L\times 4\times H'\times W'}$. In the fixed forward process, $\mathbf{z} \coloneqq \mathbf{z}_0$ is diffused into a pure noise tensor $\mathbf{z}_T$ by $T$ steps. In the reverse process (namely the denoising process), the denoiser model predicts the previous-step data $\mathbf{z}_{t-1}$ from current noisy data $\mathbf{z}_t$ by taking the embedded text prompt and frame-wise depth maps as conditions, and a clean data $\mathbf{z}'_0$ can be sampled from random noise $\mathbf{z}_T$ in a recurrent manner.
Specifically, the denoiser model is a 3D U-Net which we adopt the architecture from \cite{he2022latent}.
% comprised of two main building blocks: residual pseudo-3D convolution blocks and transformer-based attention blocks. 
We adopt CLIP~\cite{radford2021learning} as the textual encoder to extract visual-aligned tokens from the input text prompts. 
For the depth estimator, we choose the Midas depth estimation model~\cite{ranft2022towards} due to its robustness on various videos.
The depth maps are encoded through a CNN-based structure encoder and the multi-scale features are added to the feature maps of the denoiser U-Net for structural modulation. 
Different from structure control, semantic control via textual prompt affects the backbone features via a cross-attention module\cite{rombach2022high}. 
% These two conditions control the video synthesis with motion structure and visual appearance.

%---------------------------------------------
\subsection{Video Character Rerendering}
\label{subsec:coherent_synthesis}
Above mentioned video synthesis framework manages to provide videos with high-quality and diverse motion. However, the generated character appearance controlled by text prompts varies in different video clips. 
To overcome this challenge, we formulate this problem with the following objective:
Given a pre-trained video generation model, and user-specified characters, our goal is to generate consistent characters across different video clips; we referred to this task as \textit{video character rerendering}. 
To do so, we examined existing literature on personalization approaches of image diffusion models. 
However, there are several challenges associated with directly applying these methods for video personalization.
1) How to leverage image data for personalizing video models? 
One straightforward approach for video personalization is to utilize video data that portrays a specific character.
However, the video data of a consistent character is much harder to collect than images.
2) How to adjust the tradeoff between the concept compositionality and the character fidelity? This challenge also exhibits in image personalization literature.
In this section, we will explain preliminary approaches and our method in detail.
% To make the generated video clips have consistent character appearance and style, we study existing concept customization approaches and propose timestep-variable textual inversion (Time-TI) for embedding the target concept to the diffusion text embedding space. 
% Thus, the model can synthesize the target concept consistently in different clips through the guidance signal provided by the timestep-variable text embedding.
% In this section, we will explain preliminary approaches and our technical design choices in detail.

\begin{figure}[!t]
	\centering
	\includegraphics[width=1\linewidth]{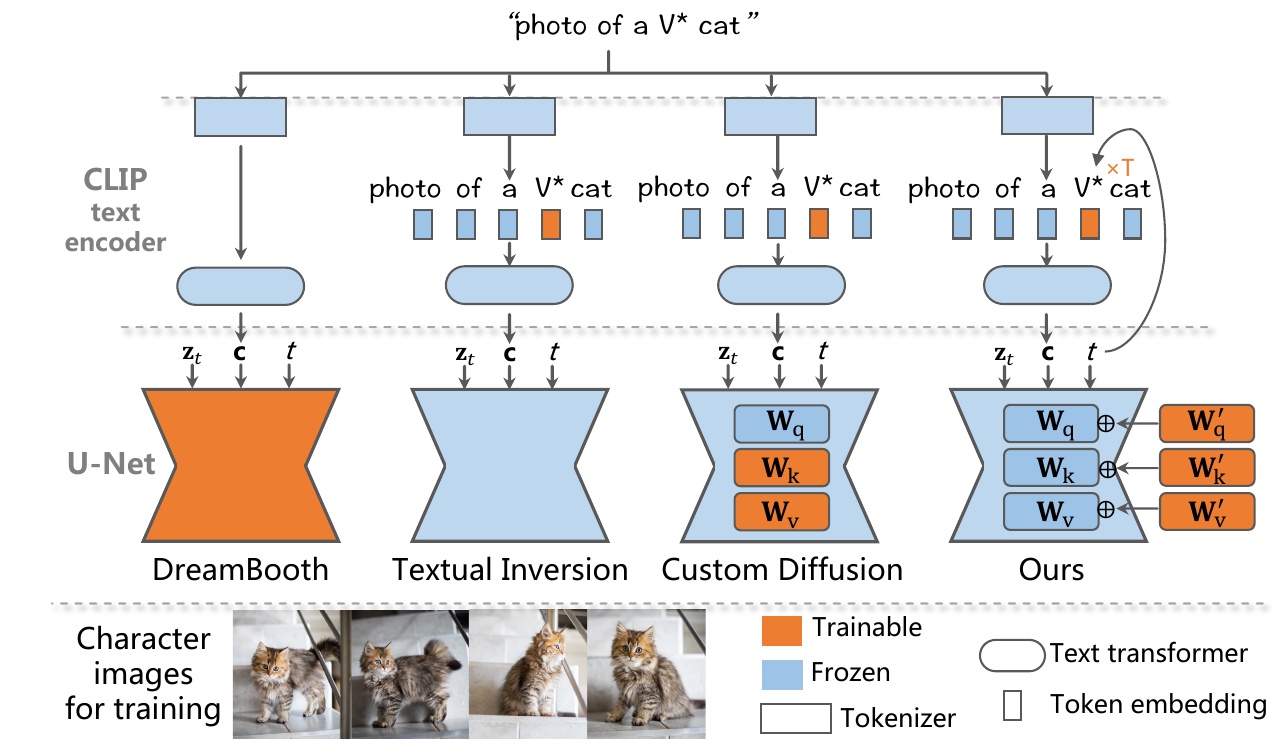}\vspace{-0.5em}
	\caption{Concept diagram of different approaches for personalization.
    To overcome the inconsistency problem of generated characters, we study existing personalization approaches and propose a new method to rerender the appearance of the target character.
    We keep all parameters from the CLIP text encoder and the denoiser U-Net frozen and learn timestep-dependent token embeddings to represent the semantic features of target characters. Additionally, we insert a new branch to the projection layers of $\mathbf{q}, \mathbf{k}, \mathbf{v}$ in attention modules and modulate the pre-trained weight to better represent the character.
    }
    \label{fig:concept_module}
\end{figure}

%---------------------------------------------
\textbf{Preliminary: Textual Inversion.}
Textual Inversion is an image personalization approach that aims to represent a new concept to a new token $\token$ and learns a corresponding new token embedding vector $\tokenembd$ in the CLIP text encoder $\clip$.
The $\tokenembd$ is directly optimized with 3-10 images depicting a specific concept.
The training objective is the same as the original loss of diffusion models, which can be defined as:
\begin{equation}
   v_* = \argmin_v \mathbb{E}_{z\sim\mathcal{E}(\mathbf{x}), \mathbf{y}, \epsilon \sim \mathcal{N}(\mathbf{0}, \mathbf{I}), t }\Big[ \Vert \epsilon - \mathbf{f}_{\theta}(\mathbf{z}_{t},t, \mathbf{c}) \Vert_{2}^{2}\Big] \, ,
    \label{eq:v_opt}
\end{equation}

After training, the new token can be combined with other word tokens to form a sentence. This token sequence can then be passed through the text encoder to obtain conditional text token embeddings that facilitate the control of image generation for producing the desired concept.

\textbf{Timestep-variable textual inversion (TimeInv).}
However, optimizing the single token embedding vector has limited expressive capacity because of its limited optimized parameter size.
In addition, using one word to describe concepts with rich visual features and details is very hard and insufficient.
Hence, it tends to suffer from unsatisfactory results regarding the fidelity of concepts. 
To tackle this problem, we propose timestep-variable textual inversion (TimeInv).
TimeInv is based on the observation that different timesteps control the rendering of different image attributes during the inference stage. 
For example, the previous timesteps of the denoising process control the global layout and object shape, and the later timesteps of the denoising process control the low-level details like texture and color~\cite{voynov2022sketch}.
% the middle timesteps control the semantics, 
To better learn the token depicting the target concept, we design a timestep-dependent token embedding table to store the controlling token embedding at all timesteps.
During training, we sample random timesteps among all ddpm timesteps to directly optimize the timestep-embedding mapping table $\mathbf{V} \in \mathbb{R}^{T\times d}$ where $T$ is the total timesteps of diffusion process and $d$ is the dimension of token embeddings.
The training objective can be defined as:
\begin{equation}
   \mathbf{V} := \argmin_{v_{1:T}} \mathbb{E}_{z\sim\mathcal{E}(x), y, \epsilon \sim \mathcal{N}(0, 1), t }\Big[ \Vert \epsilon - f_{\theta}(\mathbf{z}_{t}, t, \mathbf{c}_\theta(y, t)) \Vert_{2}^{2}\Big] \,.
    \label{eq:v_opt}
\end{equation}

During inference, the token embedding is retrieved based on the current denoising timestep and then composite to a sequence of token embeddings, defined as $v_*^t = V_t$.

\textbf{Video customization with image data.} 
Another challenge for video personalization is how to leverage image data to optimize video generation models.
Directly repeating images to videos and then optimizing the tokens will lead to the motion omission problem.
Since the static motion tends to bind with the target concept and hard to generate concepts with diverse motions.
Thanks to the previously introduced structure-guided module, now we can learn the concept using static structure guidance. 
Specifically, we repeat the concept image to a pseudo video with $L$ frames and extract framewise depth signals to control the video generation model to synthesize static concept videos.
During inference, it can easily combine the target concept with other motion guidance to generate a concept with diverse actions.

% The optimization goal becomes to \todo{figure},
% where the depth estimator and depth adapter are kept frozen during training.

\textbf{Low-rank weight modulation.}
Using textual inversion only is still hard to capture the appearance details of the given character. Instead of previous approaches that directly optimize model parameters, we add additional low-rank~\cite{hu2021lora} matrices to the pre-trained linear layers in attention modules, without hurting the concept generation and composition ability in the pre-trained model.
The low-rank matrices comprise two trainable linear layers. We insert these matrices in the cross and spatial self-attention modules in our model.

\textbf{Conflict between structure guidance and concept generation.}
Although the concept can be successfully injected into video generation with our tailored design, there still exists a severe concept-guidance conflict issue. 
% We observe another problem of this task is the concept-guidance conflict problem.
Specifically, if we want to learn a personalized teddy bear and then use a source video to provide motion guidance, it is challenging and time-consuming to collect teddy bear moving videos.
Besides, the shape provided by the depth will severely affect the id similarity because the generated shape needs to follow the id shape.
Hence it is crucial for a depth-guidance model to have the ability to relax its depth control. 
To solve this, we make our depth-guidance module to be adjustable via timestep clamping during sampling.
Concretely, we apply depth guidance on the feature only for the timesteps $t = T, ..., \tau$ and drop the depth feature after timestep $\tau$. 
We also experimented with feature rescaling during inference in early attempts, which shows worse depth adjustment than the timestep clamping.

\begin{figure}[!t]
	\centering
	\includegraphics[width=1\linewidth]{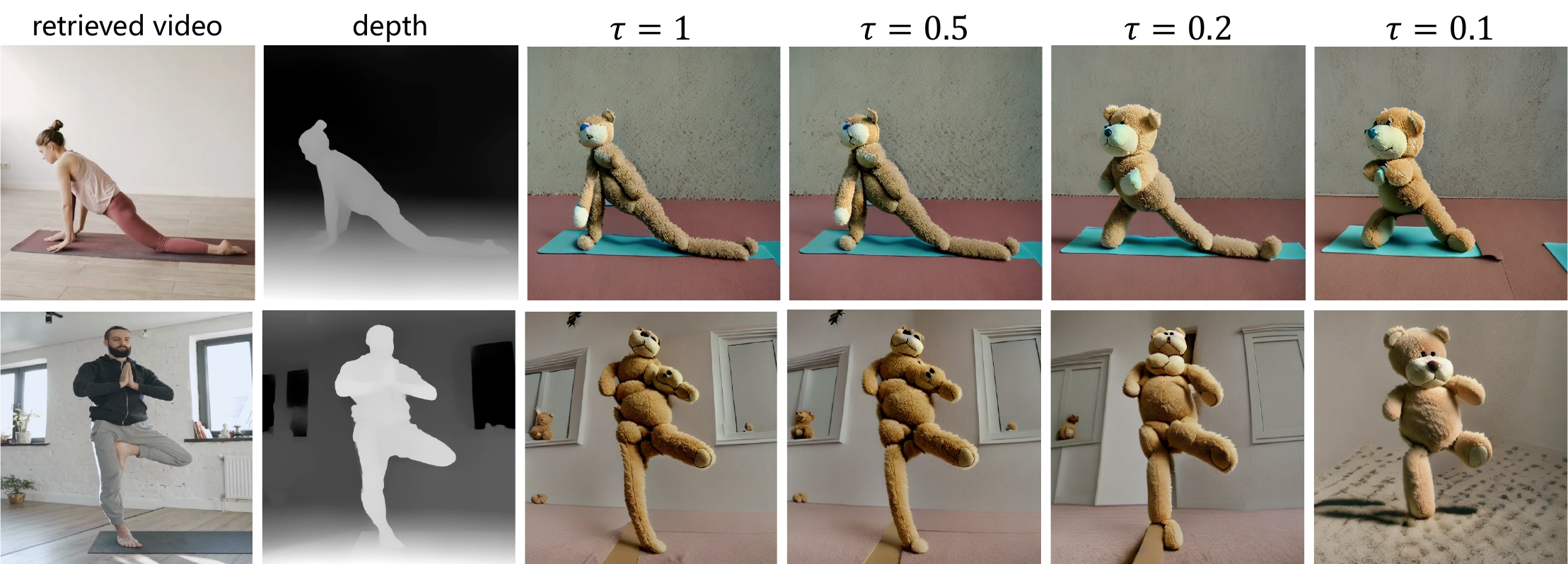}
	\caption{
    Effectiveness of adjusting the $\tau$.
    Small $\tau$ can relax the depth control to make the shape render towards the character shape while maintaining a coarse layout and action control from the depth.
    This technique can generate videos of teddy bears without the need to retrieve the motion video of teddy bears, which is very hard to collect since there is a lack of real videos of teddy bears with diverse motions (e.g., doing yoga).
    }
    \label{fig:effect_condtau}
\end{figure}

% \begin{figure*}[!t]
% 	\centering
% 	\includegraphics[width=1\linewidth]{figures/custom_compare_style.pdf}\vspace{-0.5em}
% 	\caption{Comparison results with previous customization approaches\todo{Adapter version}}
%  	\label{fig:model_overview}
% \end{figure*}

% \begin{figure*}[t] 
%     \centering
%     \includegraphics[width=\linewidth]{figures/custom_ablation.pdf}
%     \vspace{-8pt}
%     \caption{
%         Concept Customization Results for character and style-consistent video synthesis.
%         }
%     \label{fig:ablation:uncond_guide}
% \end{figure*}

\begin{figure*}[t] 
    \centering
    \includegraphics[width=\linewidth]{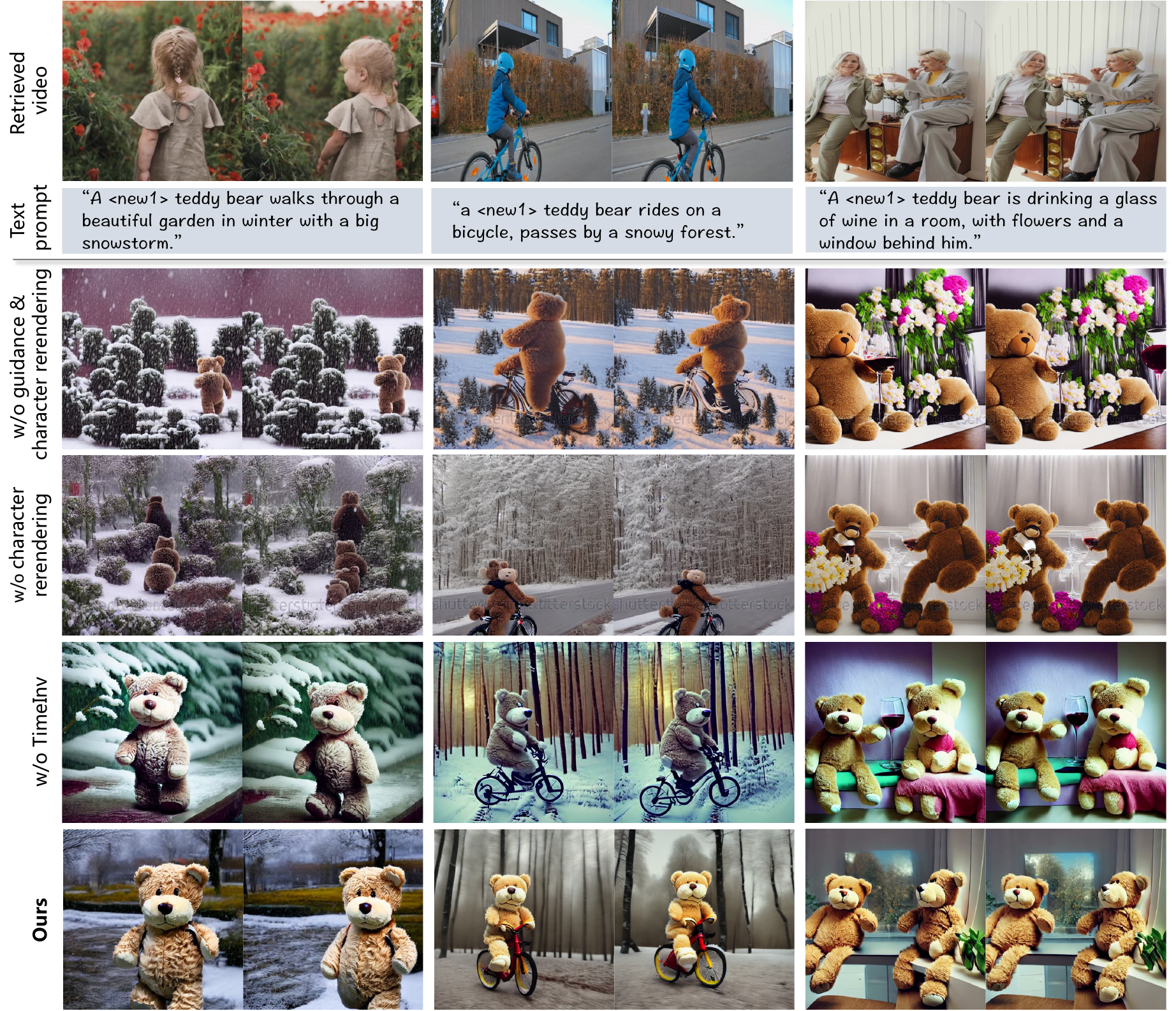}
    \caption{
        Abalation results of the core components in our pipeline, including structure guidance, character rerendering, and TimeInv.
        }
    \label{fig:compare_story2video}
\end{figure*}

\begin{figure*}[!t]
	\centering
	\includegraphics[width=1\linewidth]{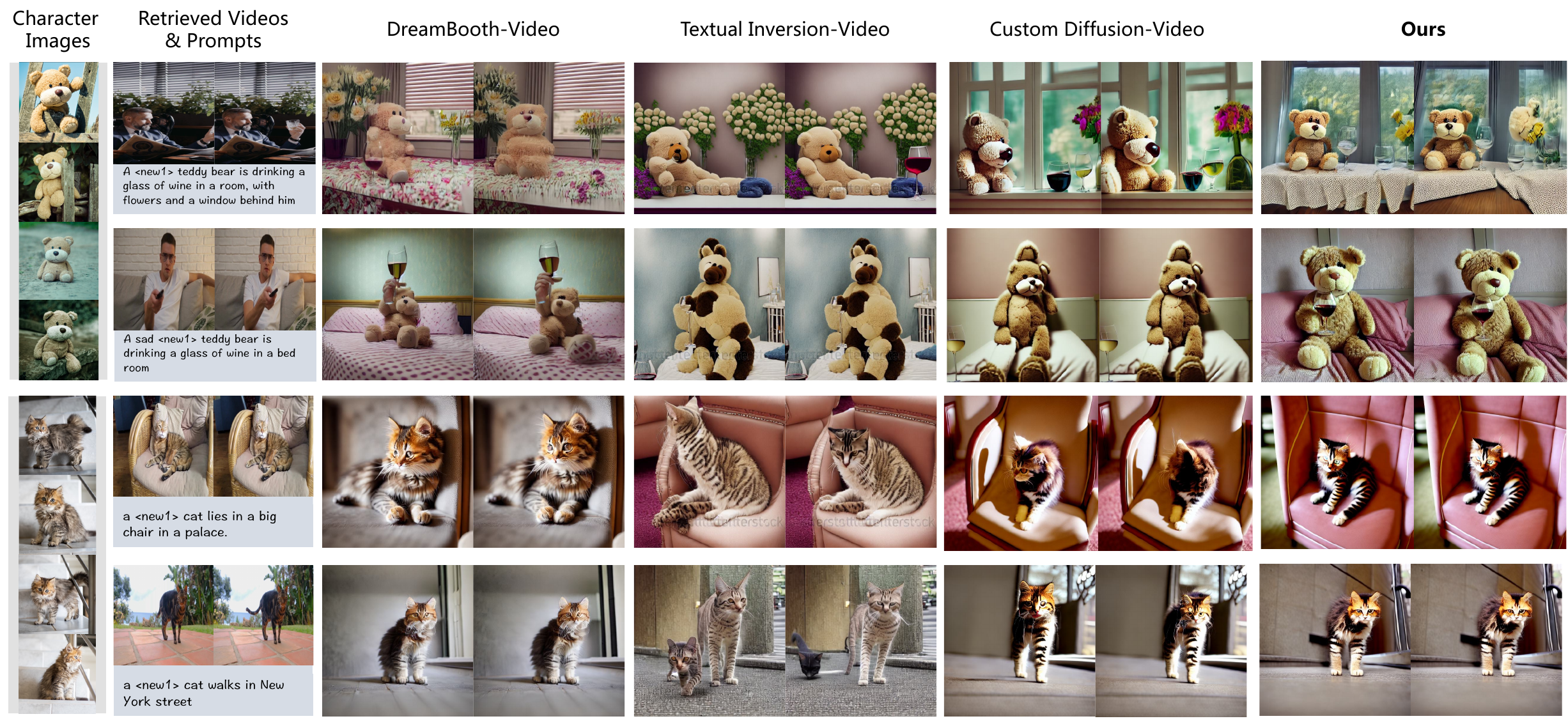}\vspace{-0.5em}
	\caption{Quantitative comparisons with previous personalization approaches. 
    We show the results of two characters using four different approaches. For each approach, we show one video clip and fix the same random seed. Each video clip shows two frames with a frame sampling stride of 8.
    Readers can zoom in for a better view.
    }
    \label{fig:compare_id}
\end{figure*}

%---------------------------------------------
\section{Experiment}
\label{sec:experiment}
%---------------------------------------------

As the storyboard splitting is conducted manually or assisted by LLMs, we mainly evaluate the retrieval-augmented video generation, as the major technical innovation of this paper. 
% As our proposed method dedicates a framework comprised of multiple technical modules, we will first assess the performance of the storyline video synthesis framework as a whole, and then evaluate the key technical modules individually.
Concretely, we first validate the effectiveness of our overall pipeline for storytelling video synthesis in \cref{subsec:eval_story2video}, and then evaluate the video synthesis quality, and the concept customization performance in \cref{subsec:eval_vidsyn} and \cref{subsec:eval_concept} respectively.

%---------------------------------------------
\subsection{Implementation Details}
\label{subsec:implementation}
%---------------------------------------------

Our video generation model is trained in three stages. Firstly, we train the base text-to-video model on WebVid-10M~\cite{brain2021frozen} dataset, with the spatial parameters initialized with the publicly available pre-trained Stable Diffusion Image LDM~\cite{rombach2022high}. WebVid-10M consists of 10.7M video-caption pairs with a total of 52K video hours. For training, we resize the videos into resolution $256 \times 256$ and sample 16 frames with a frame stride of 8. Secondly, to equip the model with depth guidance, we train the structure encoder on the same training set with the pre-trained base model frozen. At last, for concept customization, the depth-guided text-to-video model is fintuned along with specific textural tokens optimization, which is performed on task-dependent small dataset. Although our model is trained with fixed resolution, we find it support other resolutions well in inference phase.

%---------------------------------------------
\subsection{Evaluation on Storytelling Video Synthesis}
\label{subsec:eval_story2video}
%---------------------------------------------
Since we are the first work to tackle the task of storytelling video synthesis task, there are no existing baselines for comparison.
So we conduct ablation experiments to justify the effectiveness of our design choices, as shown in \cref{fig:compare_story2video}.
We can see that the generation quality of the target character deteriorates without using TimeInv.
In addition, the concept compositional ability is been damaged without TimeInv. 
For example, the model fails to generate the flowers and window in the third clip.
The absence of a personalization process makes it difficult to maintain consistency of character across various clips.
Without the inclusion of video retrieval, it becomes challenging to control the character position and the overall layout. 
Additionally, the generation performance is inferior compared to retrieval-augmented generation, as evidenced by the quantitative results presented in Table \ref{tab:eval_vidsyn}.
Video results can be checked in supplementary results.
% \todo{For visually checking our synthesized storytelling video as a whole, readers are recommended to watch our supplementary video that demonstrates several typical story2video examples in comparison with baseline methods.}

% \textbf{storyline dataset for evaluation.}

% we need to collect a storyline dataset (better to accompany ground-truth reference) and introduce its features. \textbf{3~5 stories}

% to justify the necessity of using motion structure as guidance and concept customization, we construct two baselines, i.e. ours w/o structural guidance and ours w/o concept customization, for comparison. Here we need no quantitative comparison (user study is informative as an alternative), and only need to show one or two examples for visual inspection (of course demo video is a must).

% \paragraph{quantitative comparision.}\todo{a figure}

% - GT 

% - Ours

% - Ours - w/o customization

% - Ours - w/o structure

% \paragraph{user study.}\todo{a table}
% invite users to assess our synthesized storyline videos, and design a proper questionnaire to collect informative feedback. 

%---------------------------------------------
\subsection{Evaluation on Text-to-Video Synthesis}
\label{subsec:eval_vidsyn}
%---------------------------------------------

\textbf{Baselines and evaluation metrics.}
We compare our approach with existing available (depth-guided) text-to-video models, including T2V-Zero combined with ControlNet to inject depth control~\cite{khachatryan2023text2video}, and two open-sourcing text2video generation models ModelScope~\cite{modelscope} and LVDM~\cite{he2022latent}.
We measure the video generation performance via FVD and KVD. 
The used real video dataset is UCF-101~\cite{UCF101} and we sample 2048 real videos and their corresponding class names. We use the class name as a text prompt and then sample one fake video for each prompt.
For all three models, we use DDIM 50 steps and their default classifier-free guidance scales, which are 9 for ModelScope and T2V-Zero, and 10 for LVDM.

\textbf{Results.}
In \cref{tab:eval_vidsyn}, we show quantitative results of the performance of video synthesis. 
As can be seen, equipped with depth structure guidance, text-to-video synthesis achieves significantly better performance than video synthesis from pure text.
In addition, our approach also surpasses the existing depth-guided text-to-video generation method, T2V-Zero combined with ControlNet, demonstrating the superiority of our model.

\begin{table}[t]
    \centering
    \resizebox{1.0\linewidth}{!}{
    \LARGE
    \begin{tabular}{@{}lccccc@{}}
        \toprule
        Method & Condition & FVD$\downarrow$ & KVD$\downarrow$ \\
        \toprule
        T2V-Zero + ControlNet~\cite{khachatryan2023text2video} & text + depth & 4685.27 & 168.20 \\
        \midrule
        ModelScope~\cite{modelscope} & text & 2616.06 & 2052.84 \\
        LVDM~\cite{he2022latent} & text & 917.63 & 116.63 \\
        \midrule
        \textbf{Ours}~ & text + depth & \textbf{516.15} & \textbf{47.78} \\
        % \textbf{Ours} & \todo{} & \todo{} \\
        \bottomrule
        \end{tabular}
    }
    \caption{
    Quantitative comparison with open-sourcing video generation models on UCF-101 under zero-shot setting.
    }
    \label{tab:eval_vidsyn}
\end{table}

%---------------------------------------------
\subsection{Evaluation on Personalization}
\label{subsec:eval_concept}
%---------------------------------------------
\textbf{Baselines and evaluation metrics.}
To evaluate the effectiveness of our personalization module, we compare our approach with the previous three baseline approaches: Dreambooth\cite{ruiz2023dreambooth}, Textual inversion~\cite{textualinv}, and Custom Diffusion~\cite{customdiffusion}. 
For quantitative evaluation, we measure the semantic alignment between generated videos and text and the concept fidelity between generated videos and the user-provided concept images.
For each approach, we sample 10 random videos using 20 prompt-video pairs constructing 200 generated videos in total.
The prompts range from different backgrounds and different compositions with other objects.
Then the semantic alignment is computed by the cosine similarity between the CLIP text embedding and the CLIP image embedding of each frame, then we average the scores from all frames to obtain the alignment score between a text and a generated video clip.
The concept fidelity is measured by the average alignment between generated video clips and corresponding concept images.
For each pair, we compute the cosine similarity between each frame of the clip and the target concept image, and then we average the scores from all frames.
Different approaches share the same set of random seeds for a fair comparison.

\textbf{Implementation details.}
We adopt character images from Custom Diffusion, and we experiment with a teddy bear and a cat, containing 7 and 5 images respectively.
We use <new1> as the pseudo word to represent the given character in the input text prompt.
For the character of the teddy bear, we train our approach and baseline approaches to 1,000 steps. The learning rate we used for our approach and textual inversion is 1.0e-04, while we use the learning rate of 1.0e-5 for dreambooth and custom diffusion since they directly optimize the pre-trained model parameters.
For all these approaches, we use real videos of the same category as the regularization dataset to prevent the overfitting issue. The regularization data are retrieved from the WebVid dataset. For each character, we retrieved 200 real videos.
We also use data augmentation of target character images in all approaches to enrich the diversity of the training dataset following Custom Diffusion.
During inference, we use DDIM sampling with 50 sampling timesteps and the classifier-free guidance of 15 for all approaches.

\textbf{Results.}
In \cref{fig:compare_id}, we present qualitative results of comparisons with baseline personalization approaches in the video generation setting.
As can be seen, DreamBooth updates the whole pre-trained parameters thus it tend to suffer from the overfitting problem, hindering its generation diversity (e.g., the background of the third and fourth row is very similar to the training character images).
Textual Inversion only optimizes a single token embedding to represent the target character's appearance, so it is hard to capture the character details and exhibits poor character fidelity.
Custom Diffusion updates the linear layer for computing k and v in attention modules inside the pre-trained network, combined together with textual inversion. Although it achieves better character fidelity, it frequently shows artifacts in the generated character appearance.

In \cref{tab:id_quanti_compare}, we provide quantitative results of comparisons. Our proposed TimeInv can serve as a replacement for Textual Inversion and can be combined with custom diffusion, achieving better semantic alignment.

\begin{figure}[t] 
    \centering
    \includegraphics[width=\linewidth]{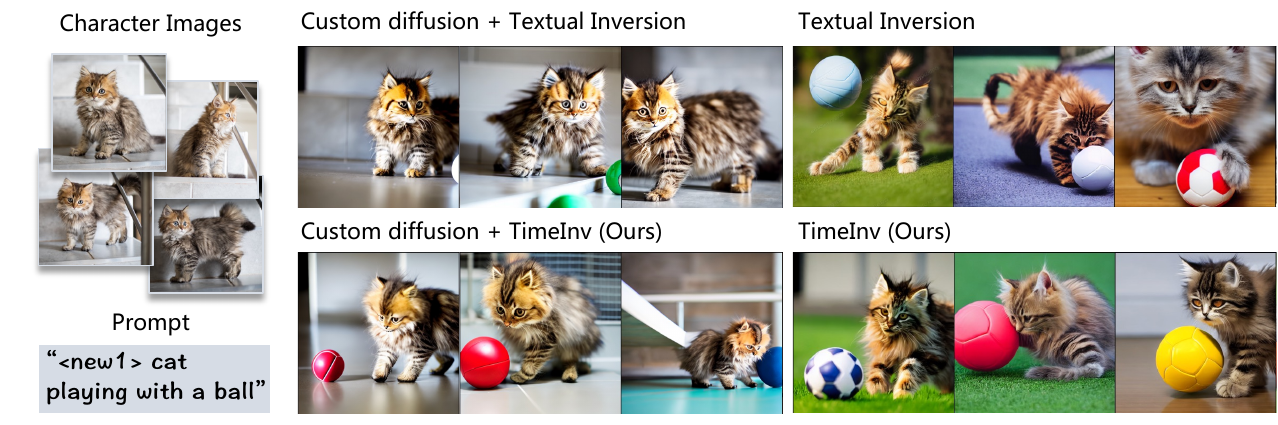}
    \caption{
        Effectiveness of the proposed Timestep-variable Textual Inversion (TimeInv) on image personalization using the pre-trained Stable Diffusion.
        Results in the same column are compared under the same training step and random seeds.
        This demonstrates that our approach can serve as a general approach for personalization on both image and video generation tasks.
        }
    \label{fig:effect_timeinv}
\end{figure}

Besides the video generation setting, we also evaluate the proposed timestep-variable textual inversion in the common image personalization setting. 
We use the Custom Diffusion codebase and compare the performance between TimeInv and Textual Inversion.
The results are shown in \cref{fig:effect_timeinv}. We can see that combining TimeInv with Custom Diffusion shows better background diversity, and concept compositionally (e.g., the ball appears more frequently than the Custom Diffusion + Textual Inversion).
Comparing TimeInv with Textual Inversion directly without updating model parameters shows that TimeInv has better character similarity (i.e., the unique texture of the cat).

% \begin{table*}[t] 
%     \centering
%     \resizebox{1.0\linewidth}{!}{
%     \begin{tabular}{@{}lcccccc}
%         \toprule
%          & \multicolumn{3}{c}{Teddy Bear} & \multicolumn{3}{c}{Cat} \\
%         \midrule
%         Method & Semantic Alignment & Temporal Consistency & ID Fidelity & Semantic Alignment & Temporal Consistency & ID Fidelity \\
%         % \toprule
%         \midrule
%         DreamBooth-Video~\cite{ruiz2023dreambooth} & 0.271 & 0.964 & 0.778 & 0.255 & 0.982 & 0.869 \\
%         Textual Inversion-Video~\cite{} & 0.263 & 0.974 & 0.665 & 0.252 & 0.968 & 0.743 \\
%         Custom diffusion-Video~\cite{} & 0.275 & 0.990 & 0.848 & 0.256 & 0.975 & 0.841 \\
%         Custom + TimeInv~\cite{} & \textbf{0.281} & \textbf{0.992} & 0.844 & 0.257 & 0.975 & 0.845 \\
%         \bottomrule
%         \end{tabular}
%     }
%     \caption{
%         Quantitative comparison with previous customization approaches.
%     }
%     \label{tab:id_quanti_compare}
% \end{table*}

\begin{table}[t] 
    \centering
    \resizebox{1.0\linewidth}{!}{
    \begin{tabular}{@{}lcccccc}
        \toprule
        \multirow{2}{*}{Method} & \multicolumn{2}{c}{Teddy Bear} & \multicolumn{2}{c}{Cat} \\
        %\midrule
        \cmidrule(lr){2-3}\cmidrule(lr){4-5} & Sem. & ID &  Sem. & ID \\
        % \toprule
        \midrule
        DreamBooth~\cite{ruiz2023dreambooth}-Video & 0.272 & 0.778 & 0.255 & 0.869 \\
        Textual Inversion~\cite{textualinv}-Video & 0.263 & 0.666 & 0.252 & 0.743 \\
        Custom diffusion~\cite{customdiffusion}-Video & 0.275 & 0.849 & 0.256 & 0.841 \\
        \textbf{Ours} & \textbf{0.295} & \textbf{0.853} & \textbf{0.257} & \textbf{0.902} \\
        %
        % Custom diffusion-Video+TimeInv & \textbf{0.281} & 0.844 & \textbf{0.257} & 0.845 \\
        \bottomrule
        \end{tabular}
    }
    \caption{
        Quantitative comparison with previous personalization approaches.
    }
    \label{tab:id_quanti_compare}
\end{table}

%         Ours & \textbf{0.277} & 0.821 & 0.252 & 0.846 \\
%         \bottomrule
        
% %---------------------------------------------
% \subsection{Limitation}
% \label{subsec:limitation}
% %---------------------------------------------

% summarize the limitation of our current system and discuss the direction for future improvement.

\section{Conclusion}
We introduce a novel retrieval-based pipeline for storytelling video synthesis.
This system enables better video synthesis quality, layout, motion control, and character personalization for producing a customized and character-consistent storytelling video.
We incorporate a structure-guided video generation module to a base text-to-video model, and we devise a new personalization method to boost the character control performance.
We also solve the character-depth confliction problem with the adjustable depth controlling module.
While we have shown the ability of our system for the challenging storyline synthesis task, there is much room for future improvement from multiple aspects. 
For example, a general character control mechanism without finetuning and a better cooperation strategy between character control and structure control can be potential directions.

% Bibliography
\bibliographystyle{ACM-Reference-Format}
\bibliography{reference}

% Appendix
% \appendix
% \input{sections/06_appendix}

\end{document}